\pdfoutput=1

\PassOptionsToPackage{dvipsnames}{xcolor}

\documentclass[sigconf, nonacm]{acmart}

\usepackage{bm}
\usepackage{nicefrac}
\usepackage{enumitem}
\usepackage{wrapfig}
\usepackage{multirow}
\usepackage{makecell}
\usepackage{subcaption}
\usepackage{float}

\AtBeginDocument{%
  }

\begin{document}

\title{Through the PRISm: Importance-Aware Scene Graphs for Semantic Image Retrieval}

\newcommand{\ntua}{%
  \affiliation{%
    \department{Artificial Intelligence and Learning Systems Laboratory}
    \institution{National Technical University of Athens}
    \city{Athens}
    \country{Greece}}}

\author{Dimitrios Georgoulopoulos}
\email{digeorgoulopoulos@gmail.com}
\ntua

\author{Nikolaos Chaidos}
\email{nchaidos@ails.ece.ntua.gr}
\ntua

\author{Angeliki Dimitriou}
\email{angelikidim@ails.ece.ntua.gr}
\ntua

\author{Giorgos Stamou}
\email{gstam@cs.ntua.gr}
\ntua

\renewcommand{\shortauthors}{Georgoulopoulos et al.}

\begin{abstract}

Web-based visual search increasingly handles queries driven by semantic intent, focusing on what is happening in an image rather than merely what it looks like. While global embeddings from vision-language models capture overall appearance, they often blur fine-grained relational structure. Conversely, scene graphs make these relationships explicit, but generating them automatically for unannotated web images produces dense, noisy graphs 
that retain every detected entity and relation regardless of relevance.
To address this, we introduce \textit{\textbf{PRISm} (\textbf{P}runing-based Image \textbf{R}etrieval via \textbf{I}mportance Prediction on \textbf{S}e\textbf{m}antic Graphs)}, an image-to-image retrieval framework that learns to focus on what matters in a scene. PRISm employs an \textbf{Importance Prediction Module}, supervised by caption-derived relevance strictly during training, to score and prune non-essential objects and $\langle\text{subject}, \text{relation}, \text{object}\rangle$ triplets. An \textbf{Edge-Aware Contextual GNN} then aggregates relation conditioned neighbor messages, fusing the refined graph with global visual features into a single, index-ready embedding per image. On PSG, PRISm outperforms the strongest baseline by $12.7$ points in MAP@1. Furthermore, on Flickr with machine-generated scene graphs and captions, PRISm remains best on top-1 metrics,
while adding only 33\,ms
per query beyond scene graph generation.

\end{abstract}



\maketitle

\section{Introduction}
\label{sec:intro}

Content-based image retrieval (CBIR) has transitioned from specialized domains to widespread deployment across web-scale platforms, including commercial engines such as Pinterest and Bing \cite{jing2015pinterest, hu2018bing}. In these environments, user intent frequently hinges on high-level semantics rather than low-level perceptual attributes; retrieval is governed not by surface-level color distributions or spatial layout, but by shared event semantics: equivalent entities engaged in similar interactions. Effectively addressing this intent requires visual representations that encode both entity presence and fine-grained relational structures, while remaining sufficiently compact to support scalable indexing and retrieval over massive web collections.

Prevailing image-to-image retrieval pipelines represent each image through
a single global embedding derived from a pretrained vision or
vision-language encoder, such as DINOv3 \cite{Simeoni2025DINOv3}, CLIP
\cite{radford2021clip} or BLIP \cite{blip}. Although such embeddings
effectively encode appearance and holistic scene gist and are inherently
amenable to efficient indexing, they remain largely insensitive to
relational structure: vision-language models have repeatedly been shown to
exhibit bag-of-words behavior, failing to discriminate between inputs that
comprise identical entities arranged in different relations
\cite{thrush2022winoground, yuksekgonul2023aro, hsieh2023sugarcrepe}. In
a retrieval setting, this deficiency manifests as matches driven by entity
co-occurrence rather than interaction semantics; a query depicting a dog
biting a skateboard, for instance, may be matched to images of people
riding skateboards, or indiscriminately to any image containing a dog
(Section~\ref{sec:qual}).

\begin{figure}[t]
    \centering
    \includegraphics[width=\linewidth]{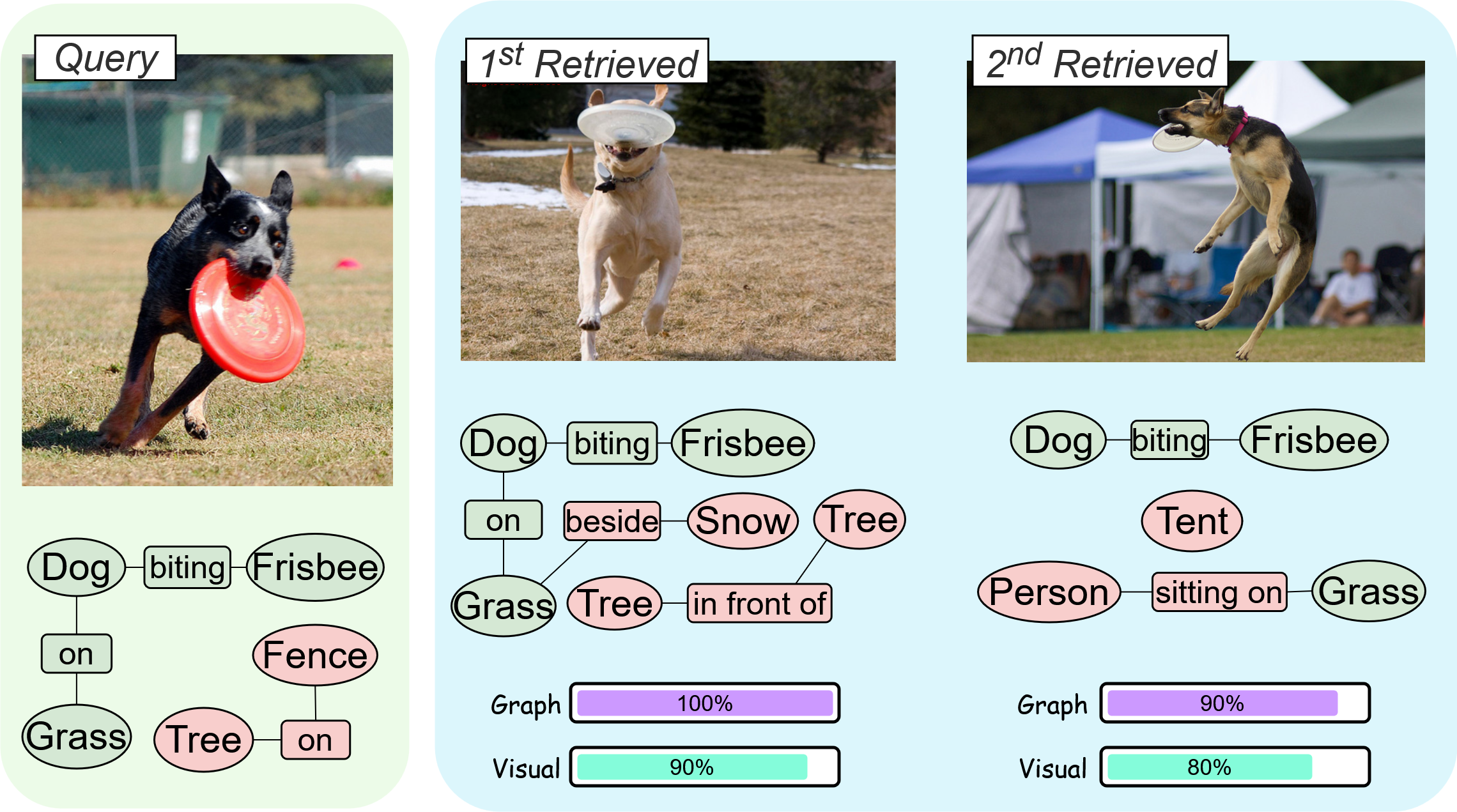}
    \caption{Example of top-2 image-to-image retrievals. PRISm (i) leverages scene graphs alongside visual features, (ii) retains important (\textcolor{ForestGreen}{green}) / prunes redundant (\textcolor{WildStrawberry}{red}) elements, and (iii) explicitly encodes object interactions, emphasizing relational structure. }
    \label{fig:teaser}
    \vspace{-10pt}
\end{figure}

Scene graphs (SGs) constitute an explicit, structured alternative \cite{johnson2015image, krishna2016visualgenome}. By encoding objects as nodes and their interactions as $\langle$subject, relation, object$\rangle$ triplets, SGs expose precisely the relational structure that global embeddings obscure, and a growing body of SG-based retrieval methods learns to assess image similarity through graph-level comparison \cite{yoon2021imagegraph, wang2023hisigir, chaidos2025scenir}. Nevertheless, two fundamental obstacles impede the transfer of this paradigm to web-scale image search. First, existing approaches treat all objects and relations within an SG uniformly, notwithstanding that only a small subset conveys the core semantics of an image: human descriptions of a scene consistently reference only a fraction of the depicted entities \cite{spain2011importance, berg2012importance}, and even curated datasets such as Visual Genome \cite{krishna2016visualgenome} and Panoptic Scene Graph (PSG) \cite{yang2022panoptic} contain numerous nodes and edges tangential to the principal semantics, which introduce noise into the learned representations and may mislead retrieval \cite{chaidos2025scenir, dimitriou2024structure}. Second, web imagery is inherently devoid of SG annotations; structure-aware retrieval must therefore rely on scene graph generators, whose outputs are noisy and biased toward frequent predicates \cite{tang2020unbiased}. Empirically, graphs generated for Flickr30k images contain approximately twice as many relations per image as the human-annotated PSG graphs, despite a comparable number of objects (Table \ref{tab:psg_flickr_stats}). Robustness to noisy structure thus constitutes a prerequisite for the semantic mining of multimedia web content, and forms the central challenge addressed in this work.

Figure \ref{fig:teaser} exemplifies this challenge. The query image comprises three key entities - dog, frisbee and grass - whose interactions, $\langle$dog, biting, frisbee$\rangle$ and $\langle$dog, on, grass$\rangle$, largely determine the interpretation of the scene. In contrast, the fence, the tree and the relation linking them constitute peripheral background content. The scene graph, however, retains all of
these elements indiscriminately; a retriever that consumes the complete
graph therefore inevitably devotes part of its representational capacity
to matching semantically irrelevant content. Current pipelines rely either
on global encoders lacking any explicit notion of objects and relations
\cite{blip, radford2021clip, Simeoni2025DINOv3} or on complete, unfiltered
scene graphs \cite{chaidos2025scenir, wang2023hisigir}; neither provides a
mechanism to discard redundant information in
both human-annotated and generated SGs.

To address these limitations, we introduce \textbf{PRISm} (\textbf{P}runing-based Image \textbf{R}etrieval via \textbf{I}mportance Prediction on \textbf{S}e\textbf{m}antic Graphs), a multimodal framework for importance-aware scene graph retrieval. Its first component, the \textbf{Importance Prediction Module} (IPM), estimates the semantic importance of each object and $\langle$subject, relation, object$\rangle$ triplet from joint visual and textual cues, and prunes the graph to its most salient elements. Human-written captions serve as a proxy for semantic relevance exclusively during training; at inference, the IPM operates solely on the image and its graph, rendering it directly applicable to unannotated web images and their generated SGs. In Figure~\ref{fig:teaser}, the IPM consistently retains $\langle$dog, biting, frisbee$\rangle$ and the grass node across graphs, while pruning other background objects and their interactions. By grounding retrieval in these core elements, PRISm aims to mitigate the dependence of retrieval quality on graph quality.

The pruned graph is then encoded and fused with global visual context. In standard attention-based GNNs, a neighboring node transmits an identical message irrespective of its relation to the target node; a ``person'' node, for instance, contributes equivalently whether it is \emph{playing with} or \emph{sitting next to} the target. The proposed \textbf{Edge-Aware Contextual GNN} instead conditions each neighbor representation on its specific relation to the target node prior to attention computation, such that relational information governs both the attention distribution and the content of the propagated messages. Overall, PRISm integrates three feature types (the pruned SG, object-level visual features, a global image embedding) within a dual-stream architecture: a \emph{Graph Stream}, which applies the Edge-Aware GNN to the pruned SG with textual and visual node attributes, and a \emph{Global Visual Stream}, which captures holistic scene context. The outputs of both streams are concatenated into a single compact vector per image, enabling candidate embeddings to be precomputed and queried via standard indexing. As shown in Figure~\ref{fig:teaser}, the top-ranked retrieval not only preserves $\langle$dog, on, grass$\rangle$, which is absent from the second result, where the grass participates only in the pruned $\langle$person, sitting on, grass$\rangle$ triplet, but also exhibits closer visual correspondence to the query in terms of the dog's pose, orientation and framing. This synergy between semantic structure and visual context yields retrievals more closely aligned with caption-derived semantic similarity.

In summary, the contributions of this work are: \textbf{(a)} An \textbf{Importance Prediction Module} that prunes SGs to their
  semantically salient objects and triplets, learned from caption-derived
  supervision yet caption-free at inference, and an \textbf{Edge-Aware
  Contextual GNN} that conditions message passing on relation semantics; \textbf{(b)} \textbf{PRISm}, a dual-stream framework that unifies both modules
  with global visual context into a single index-ready embedding per
  image, supporting sub-second query processing, together with a
  comprehensive evaluation on both curated and fully machine-generated
  scene graphs.
Overall, PRISm delivers the best top-ranked retrieval performance on curated \emph{and} generated SGs. On Flickr, where both SGs and captions are machine-generated, 
PRISm's own components add only 33\,ms per query, with scene graph
generation accounting for about 94\% of the $\sim$0.5\,s total; and
unlike matching-based SG retrievers, whose per-query cost grows linearly
with the gallery, its single-vector embeddings can be pre-indexed for
sublinear search.
\section{Related Work}

\textbf{Web-Scale Visual Search.}
Visual search systems deployed on the Web represent an image with a dense embedding and serve queries
through approximate search over Web-scale collections
\cite{jing2015pinterest, hu2018bing}. Contemporary pipelines derive such
embeddings from large pretrained encoders, either vision backbones such as
DINOv3 \cite{Simeoni2025DINOv3} and MambaVision
\cite{Hatamizadeh2024MambaVisionAH}, or vision-language (VL) models such as
CLIP \cite{radford2021clip}, BLIP \cite{blip} and BLIP-2 \cite{blip2},
which constitute strong, general-purpose baselines for image-to-image
retrieval. Nevertheless, a growing body of evidence indicates that these
representations encode compositional structure only weakly:
VL models perform close to chance on benchmarks requiring the
discrimination of identical entities in different relations
\cite{thrush2022winoground}, exhibit bag-of-words behavior with respect
to relational and attribute binding \cite{yuksekgonul2023aro}, and
continue to struggle once language-prior shortcuts are removed from such
benchmarks \cite{hsieh2023sugarcrepe}. PRISm preserves the single-vector,
index-compatible form of these representations, while complementing
global visual context with explicit, filtered relational structure.

\textbf{Scene-graph based Image Retrieval.} 
Following advances in graph similarity learning \cite{Ling2020graphsim1, 
Zhu2020UnsupervisedGraphSim, Zhuo2024GraphSimSupervised}, we adopt a similar perspective for SG-based image retrieval.
Unlike molecular or citation graphs, whose structure is directly observed, SGs are inferred from visual content: they are multimodal, grounded in image regions, and encode rich yet often ambiguous relations. When
extracted automatically from web imagery, they are also noisy and dense. We therefore adapt the underlying graph module to
better exploit the relational information specific to visual scenes.
Since the seminal work of
Johnson et al. \cite{johnson2015image}, several strategies have been
explored for SG-based image retrieval. For instance, IRSGS \cite{yoon2021imagegraph} and SGCE \cite{dimitriou2024structure} proposed a siamese GNN architecture trained with various surrogate similarity scores as supervision to learn global graph embeddings. 
SCENIR \cite{chaidos2025scenir} instead introduced an unsupervised Graph Autoencoder framework for substantial gains in efficiency, while \citet{Maheshwari2021contrastiveSG} implemented a contrastive-based triplet loss approach on a traditional GNN backbone, enhanced with custom probabilistic triplet sampling. 
Hi-SIGIR \cite{wang2023hisigir} addressed the lack of visual cues by fusing them with semantic features and computing local and global similarity scores. Being a matching network, it scores each query-candidate pair jointly, so candidates cannot be pre-indexed in a vector search engine and the per-query cost grows linearly with the size of the gallery.
Beyond this, these approaches share two limitations: their
GNNs often do not natively incorporate edge features during message
passing, and they operate on the complete scene graph, retaining every
object and relation irrespective of its relevance. The former limits how
faithfully graph similarity reflects semantic relevance, while the latter
is exacerbated in web settings, where SGs are obtained from generators whose outputs are noisy and biased toward frequent
predicates \cite{tang2020unbiased, li2024scene}. We address these
limitations via our proposed Edge-Aware Contextual GNN, through
contextualized latent representations for each neighbor, and the
Importance Prediction Module, which prunes redundant and tangential
elements; unlike matching-based retrieval, PRISm maps each image to a single, index-ready vector.

\textbf{Importance Estimation and Relation-Aware Graph Learning.}
Relation importance has been studied for SG generation, where
relationships are ranked by caption-derived salience
\cite{wang2020sketching}, rather than for retrieval. Graph structure
learning methods refine topology through edge re-weighting or pooling
\cite{graphweight2, graphpooling2}, yet are driven by topological or
task-loss signals and lack cross-modal semantic grounding, rendering them
not directly applicable as baselines. Pre-existing GNNs that account for edge-information incorporate
relations through fixed composition operators \cite{vashishth2020compgcn},
additive edge features \cite{hu2020gine} or joint edge updates
\cite{wang2021egat}. 
In contrast, PRISm prunes graphs according to semantic relevance
learned from caption-derived supervision, while our Edge-Aware Contextual GNN dynamically conditions neighbor messages on their specific relational interaction before computing attention.

\section{Method}

This section provides details on our framework, 
\textbf{PRISm} (\textit{\textbf{P}runing-based Image \textbf{R}etrieval via \textbf{I}mportance-prediction on \textbf{S}e\textbf{m}antic Graphs}), outlining the task, architecture, submodules, and key design choices.

\subsection{Notation and Problem Formulation}

We begin by introducing key notation. This paper addresses the task of image-to-image retrieval by leveraging the rich semantic structure of SGs. Our dataset consists of pairs $(I, G)$, where $I$ is an image and $G$ is its corresponding scene graph. Formally, a SG is defined as $G = (V, E)$, where $V$ is a set of nodes representing depicted objects, and $E \subseteq V \times V$ is a set of edges representing the relationships between them. Node information is usually formulated as a \textit{feature matrix} $\mathbf{X}\in \mathbb{R}^{n\times d}$, where $n=|V|$ and $d$ is the feature dimensionality. Similarly, an \textit{adjacency matrix} $\mathbf{A}\in \mathbb{R}^{n\times n}$, represents the edge information in a structured manner. For the task of SG-based image-to-image retrieval, the framework takes a query image/SG pair $(I_q, G_q)$, and a set of candidate pairs $\{(I_{k_1}, G_{k_1}),...,(I_{k_n}, G_{k_n})\}$. Its goal is to produce a permutation of the candidates, ranked to approximate a ground-truth surrogate similarity score $sim(I_q,I_{k_i})$ defined 
on the images.

To fully utilize information from all available modalities (image $I_q$, textual descriptions of objects/relations, scene graph $G_q$), we design custom \textbf{GNN} modules, and leverage image and text encoders, to merge different inputs into a single unified representation.

\subsection{Proposed Model}

\textbf{PRISm} is a robust multimodal framework for image-to-image retrieval using SGs. The main architecture comprises two distinct sequentially arranged stages: (i) the \textit{Importance Prediction and Pruning stage}: potentially irrelevant semantic information is detected and pruned, and (ii) the \textit{Multimodal GNN-based Retrieval} stage: fused graphs with visual and textual information get processed by a contextual Edge-Aware GNN to retrieve similarity rankings.  

\subsubsection{Importance Prediction Module}

The Importance Prediction Module (IPM) produces more \textbf{informative and focused SGs} by identifying the most semantically relevant objects and relations in an image. 
We achieve this by predicting the importance of both individual \textbf{objects} and interconnected \textbf{triplets},
which are formulated as tuples $t_i=(o_a,r_{ab},o_b)$ that contain two objects and their relationship. 
The key intuition in defining a ground-truth importance score is that human-written captions provide a strong proxy for an image's semantic content, a standard established protocol in SG-based retrieval \cite{yoon2021imagegraph, wang2023hisigir}. To mitigate the limitations and biases of any single description, we leverage five captions per image, averaging similarities to obtain a robust estimate of semantic relevance that better reflects the underlying scene. 
Therefore, the importance score is defined based on how strongly an object's or triplet's textual description aligns with the image's manual captions $C=\{c_1,...,c_K\}$ in the embedding space.

\begin{figure}[t]
    \centering
    \includegraphics[width=0.45\textwidth]{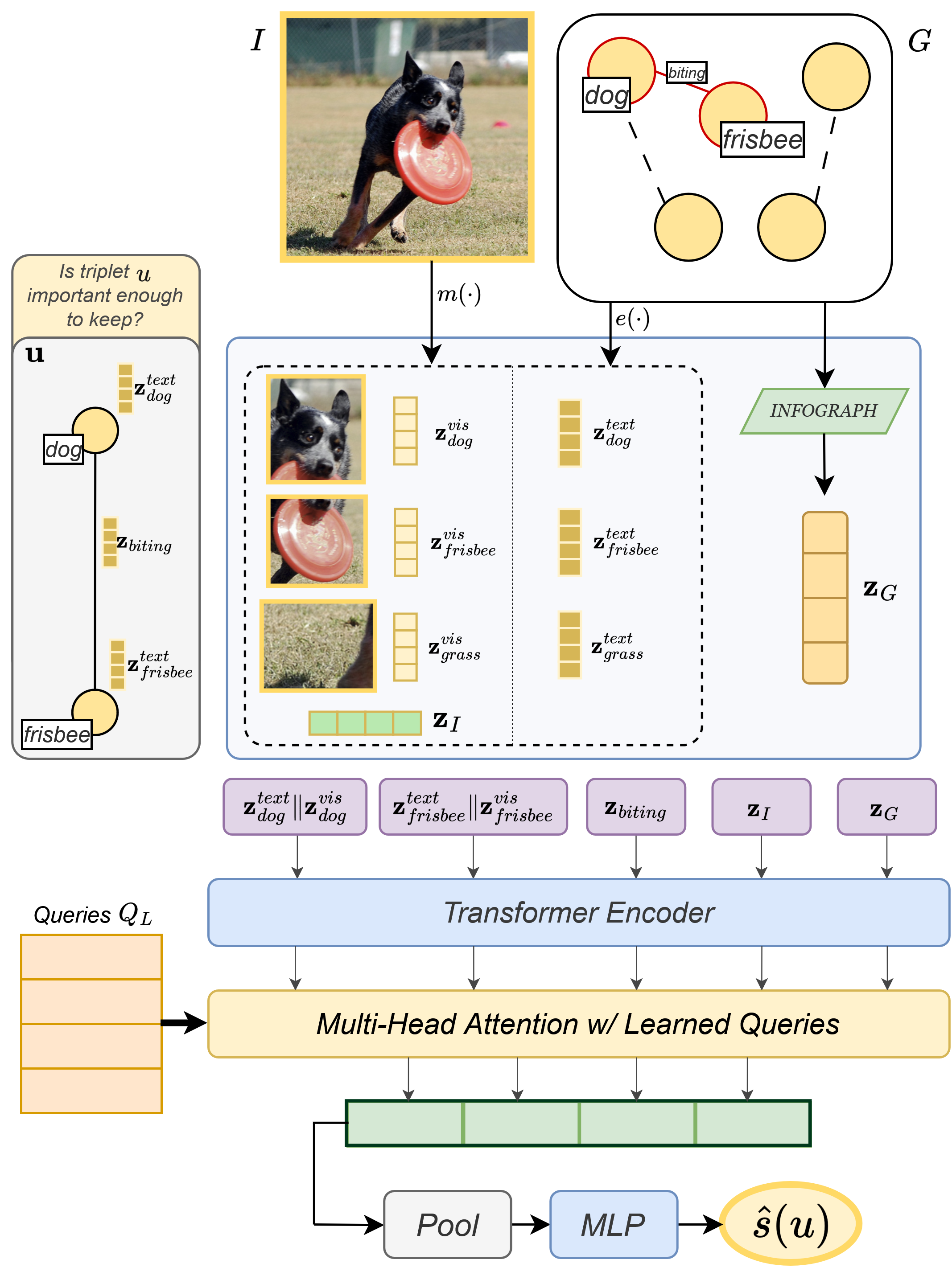}
    \caption{\textbf{Importance Prediction Module inference.} 
    Extracted embeddings of image/SG, 
    depicted objects 
    and object/relation labels
    are fed to the \textit{trained} module to predict scores $\hat{s}(u)$ and retain key objects and triplets. }
    \label{fig:importance_prediction_module}
    \vspace{-10pt}
\end{figure}

For an object $o_a$, the importance score $s(o_a)$ is defined as:
\begin{equation}
s(o_a) = \frac{1}{K}\sum_{k=1}^{K} \left\langle e(o_a),\, e(c_k) \right\rangle
\end{equation}
As for a triplet $t = (o_a, r_{ab}, o_b)$, we represent it as a space-separated phrase
$p(t)=\text{``}o_a\ r_{ab}\ o_b\text{''}$ and compute $s(t)$ as:
\begin{equation}
s(t) = \frac{1}{K}\sum_{k=1}^{K} \left\langle e\left(p(t)\right),\, e(c_k) \right\rangle
\end{equation}
where $e(\cdot)$ is a frozen sentence embedding model, to map textual input to a vector space. Similarly, we utilize a frozen vision encoding model $m(\cdot)$ to encode visual information. For score prediction, a \textbf{Transformer-Based Scoring Module} (Figure \ref{fig:importance_prediction_module}) is used, as its core self-attention mechanism is ideally suited to model the complex interactions and compute the relative importance among a pre-determined set of input tokens. For each triplet $t_i$, these input tokens are:

\begin{description}[topsep=2pt, itemsep=2pt, leftmargin=2.2cm, labelindent=0.4cm]
    \item[{$\mathbf{z}_a = \left[\mathbf{z}_a^{text} \Vert \mathbf{z}_a^{vis}\right]$}] feature vector for object $a$

    \item[{$\mathbf{z}_b = \left[\mathbf{z}_b^{text} \Vert \mathbf{z}_b^{vis}\right]$}] feature vector for object $b$

    \item[$\mathbf{z}_{ab}$] Relation feature vector

    \item[$\mathbf{z}_I$] Global visual embedding of the entire image

    \item[$\mathbf{z}_G$] Global semantic embedding of the SG
\end{description}
where $\mathbf{z}_a^{text}$, $\mathbf{z}_b^{text}$ and $\mathbf{z}_{ab}$ are generated with encoder $e$, and $\mathbf{z}_a^{vis}$, $\mathbf{z}_b^{vis}$ and $\mathbf{z}_I$ are generated with encoder $m$ (on the cropped bounding boxes, for the objects' embeddings). The global semantic embedding $\mathbf{z}_G$ is obtained by applying the unsupervised \textsc{InfoGraph} algorithm \cite{sun2020infograph} to a text-only 
version of the input SG, in order to explicitly capture the structure and the semantics of the image (more details in Appendix \ref{appendix-infograph}). 
For single-node importance scores, we set $\mathbf{z}_b = \mathbf{z}_{ab} = \mathbf{0}$ to focus on the target object.

These input features, collectively denoted $\mathbf{Z}_{u}$, are first projected to a common dimension, and passed through a
Transformer Encoder to produce contextualized embeddings $\mathbf{Z}_{enc}$. 
To aggregate the contextualized token representations into a fixed-size representation without resorting to static pooling, we adopt a query-based pooling mechanism following \citet{arar2022learnedqueries}. Specifically, we introduce a set of $K$ learnable query vectors, $Q_L = \{\mathbf{q}_1, \dots, \mathbf{q}_K\}$, which attend over the encoder outputs via Multi-Head Attention to extract a compact set of summary embeddings, $\mathbf{H}_{s}\in \mathbb{R}^{K\times d}$:
\begin{equation}
\begin{aligned}
\mathbf{Z}_{enc} &= \text{TransformerEncoder}(\text{MLP}(\mathbf{Z}_u)) \\
\mathbf{H}_{s} &= \text{MHA-LQ}(Q_L, \mathbf{Z}_{enc}, \mathbf{Z}_{enc})
\end{aligned}
\end{equation}
Embeddings $\mathbf{H}_{s}$ are mean-pooled and fed to an MLP, producing the final $\hat{s}(u)$ for a given triplet or object (Fig. \ref{fig:importance_prediction_module}).
We train this module independently, with an MSE loss between predictions $\hat{s}(u)$ and ground-truth scores $s(u)$:
\begin{equation}
\mathcal{L}
= \frac{1}{|\mathcal{S}|}\sum_{u \in \mathcal{S}} \big(\hat{s}(u) - s(u)\big)^2 ,
\end{equation}
where $\mathcal{S} = \mathcal{O} \cup \mathcal{T}$ is the set of all extracted objects and triplets from the train set.

Once the scores $\hat{s}(u)$ are either computed (for training) or predicted (for inference/test), we use them to prune the SG. In this way, we filter out less important objects and triplets, keeping the most salient information for the final retrieval. We employ three simple rules to identify retained items:

\begin{enumerate}[topsep=2pt, itemsep=2pt, leftmargin=1cm, labelindent=0.4cm]
    \item \textbf{Absolute: }
    Items with score above a fixed threshold $b$.

    \item \textbf{Relative: }
    Items designated in the high-valued class by the unsupervised \textit{Jenks Natural Breaks} \cite{jenks1967data} method with two clusters. (details in Appendix \ref{appendix-jenks}).

    \item 
    Objects that are part of any retained triplet, regardless of the previous rules.
\end{enumerate}

If any single rule of the above holds true, the object/triplet is deemed important. The final filtered graph $G^{p}$ is the induced subgraph of $G$ over the set of important objects and relations, reducing complexity and 
grounding the GNN in key semantics.

\subsubsection{Edge-aware Contextual GNN}

In our SGs, edges represent explicit textual relations (e.g., ``person - playing with - dog''). Discarding this information, as in standard GNNs like GAT or GIN, or using it only for a final matching score, is a missed opportunity. 

We propose an \textbf{Edge-Aware Contextual GNN layer} based on GATv2 \cite{brody2022gatv2} that injects relational information directly into the message-passing and attention mechanisms. The core intuition is to create \textbf{contextualized target-specific embeddings} before attention is ever computed. Crucially, this means a neighbor's representation is dynamically conditioned based on its \textit{specific relationship} to the target node, rather than being a single static embedding. Instead of just \textit{"person"}, the message from a neighbor becomes \textit{"person (who is playing with...)"}, a representation already aware of its specific relationship to the target node.

During message passing to a target node $i$ from $j$, we create a contextualized textual embedding $\tilde{\mathbf{z}}^{\text{text}}_{i\leftarrow j}$. We fuse the neighbor's textual feature $\mathbf{z}^{\text{text}}_j$ with the connecting edge feature $\mathbf{z}_{ij}$ (textual embedding of the relation) using an MLP:
\begin{equation}
\tilde{\mathbf{z}}^{\text{text}}_{i\leftarrow j} = \mathrm{MLP}\!\left(\big[\;\mathbf{z}^{\text{text}}_j \;\Vert\; \mathbf{z}_{ij}\;\big]\right)
\end{equation}
This MLP learns a joint
textual
representation 
which is then re-combined with the neighbor's original visual feature to form the full contextualized neighbor embedding, $\mathbf{z}_{i\leftarrow j}$:
\begin{equation}
    \mathbf{z}_{i\leftarrow j} = \big[\;\tilde{\mathbf{z}}^{\text{text}}_{i\leftarrow j} \;\Vert\; \mathbf{z}^{\text{vis}}_j\;\big]
\end{equation}
Only after this contextualization do we apply a variation of the standard attention mechanism to update the node embeddings. The key difference lies in how we compute the attention logits $e_{ij}$ between the target node $\mathbf{z}_i$ and the new relation-aware neighbor embedding $\mathbf{z}_{i\leftarrow j}$:
\begin{equation}\label{eq:attn_logit}
e_{ij} = \mathbf{a}^{\top}\,\mathrm{LeakyReLU}\!\left(\mathbf{W}_{\mathrm{att}}\,\big[\;\mathbf{z}_i \;\Vert\; \mathbf{z}_{i\leftarrow j}\;\big]\right)
\end{equation}
Using the now contextualized logits $e_{ij}$, the attention weights and node update for the next layer embedding $\mathbf{z}_{i, l+1}$ for node $i$, follows \citet{brody2022gatv2}.
This edge-aware mechanism is applied \textbf{only to the first layer}, a deliberate design choice, since the first layer's role is to inject edge information into the dynamic node representations. Subsequent layers can then propagate and refine these relation-aware node embeddings throughout the SG, without accidentally biasing them with over-exposure to the relations. This is both effective and efficient, as it avoids updating edge embeddings (Figure \ref{fig:retrieval}). \looseness=-1

\subsubsection{PRISm: Architecture and Training}

\begin{figure}[t]
    \centering
    \includegraphics[width=\linewidth]{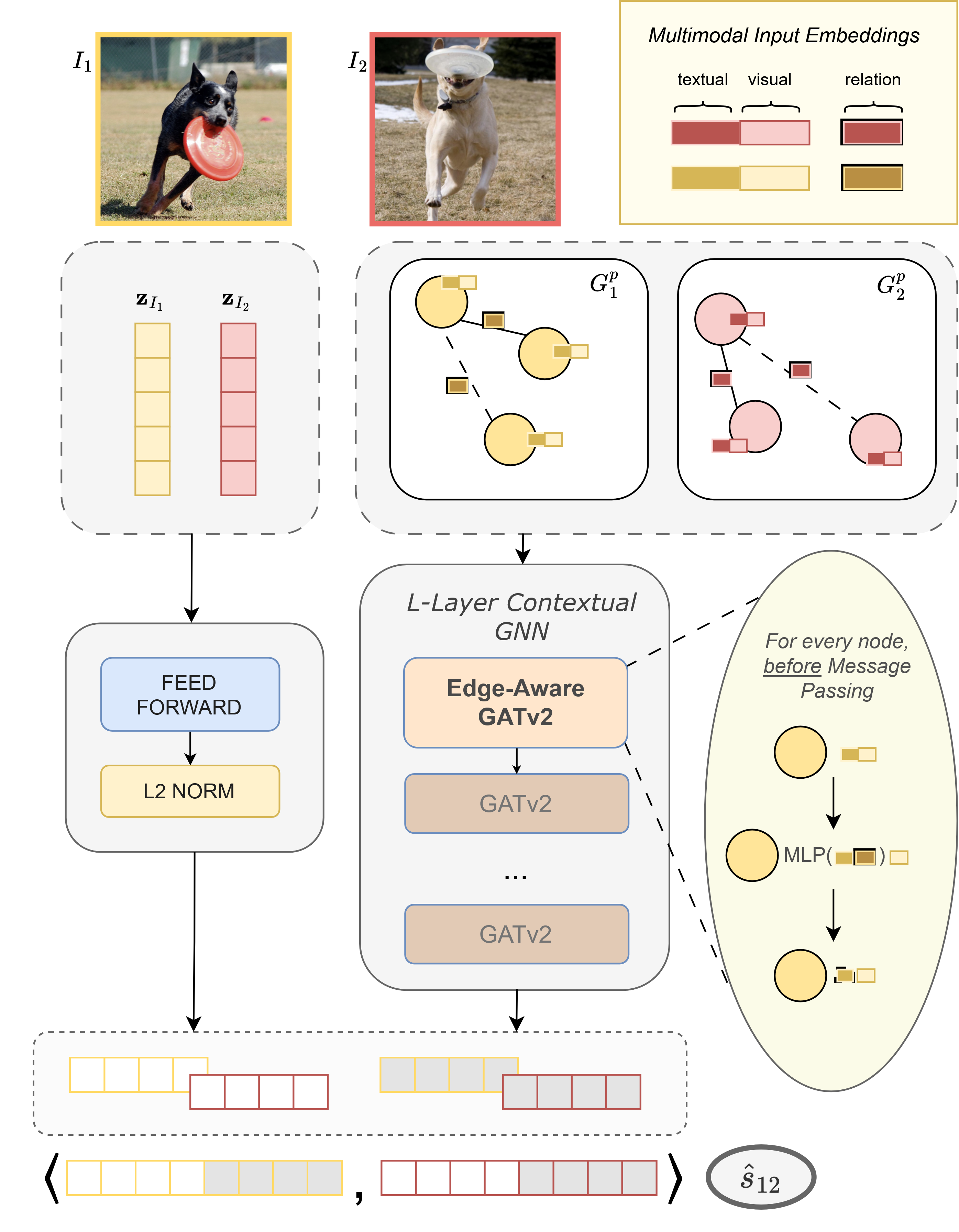}
    \caption{\textbf{Overall retrieval pipeline of PRISm.} 
Similarity between images $I_1$ and $I_2$ is computed as the inner product of embeddings that combine projected global visual features with edge-aware multimodal scene graph representations.}
    \label{fig:retrieval}
    \vspace{-10pt}
\end{figure}

The final goal of PRISm is to produce a comprehensive, multimodal embedding for each image, enabling fine-grained retrieval. Our architecture, shown in Figure \ref{fig:retrieval}, achieves this by fusing two distinct but complementary information sources: (1) a \textit{global} visual embedding that captures the overall scene context, and (2) a \textit{local} semantic graph embedding derived from our Edge-Aware GNN.

\textbf{Multimodal Graph Construction.}
Before GNN processing, the pruned $G^p$ is constructed as a multimodal entity.

\begin{itemize}[topsep=2pt, itemsep=2pt, leftmargin=0.7cm]
    \item \textbf{Nodes:} Each node $u$ (object) is represented by $\mathbf{z}_u = [ \mathbf{z}_u^{\text{text}} \parallel \mathbf{z}_u^{\text{vis}} \parallel \mathbf{b}_u ]$, concatenating its textual embedding, visual features from the object crop, and normalized spatial vector $\mathbf{b}_u = [\frac{x_1}{W}, \frac{y_1}{H}, \frac{x_2}{W}, \frac{y_2}{H}, \frac{A_u}{A_{\text{I}}}]$, where $x_{\cdot}, y_{\cdot}$ denote bounding box coordinates and $A_u$ the crop area.
    \item \textbf{Edges:} Each edge $\left(u, v\right)$ (relation) is represented by its textual embedding $\mathbf{z}_{uv}$.
\end{itemize}

\textbf{Dual-Stream Processing.}
The PRISm pipeline processes these inputs in two parallel streams:

\begin{enumerate}[topsep=2pt, itemsep=2pt, leftmargin=1cm, labelindent=0.4cm]
    \item \textbf{Global Visual Stream:} A shallow MLP processes the visual embedding of the entire image $\mathbf{z}_I$. The output, normalized to have an 
    L2-norm of $\alpha$ (tunable hyperparameter), constitutes the final visual embedding $E_I$.
    \item \textbf{Graph Stream:} Our $L$-layer Edge-Aware Contextual GNN processes $G^p$ producing a graph embedding $E_G$.
\end{enumerate}
The final multimodal representation of image $I_i$, $E^M_i$, is the concatenation of these two embeddings.

\textbf{Normalizing} of the global visual stream is crucial for balancing the multiple modalities. Capping the magnitude of $E_I$ prevents it from dominating the final representation, forcing the model to use global features as a \textit{coarse two-stage retrieval filter}, finding broadly similar images while relying on the fine-grained semantic and local information from $E_G$ for the final subtle distinctions.

\textbf{Training Objective.}
We formulate the training objective as a regression task, supervised by the ground-truth semantic similarity $s_{ij}$ between two images, $I_i$ and $I_j$. The inner product of the final embeddings, $\hat{s}_{ij} = \langle E^M_i, E^M_j \rangle$, approximates the surrogate $s_{ij}$ (defined as the average pairwise similarity between the caption embeddings of $I_i$ and $I_j$),
following previous work \cite{yoon2021imagegraph, wang2023hisigir}, which aims to align as much as possible with human perception of similarity:
\begin{equation}
    s_{ij} = \frac{1}{|C_i||C_j|} \sum_{n \in C_i} \sum_{m \in C_j} \langle n, m \rangle   
\end{equation}
To force the model to excel at distinguishing highly similar images (crucial for retrieval), we use a weighted regression loss. The contribution of each pair $(i, j)$ to the total loss is weighted exponentially by its similarity:
\begin{equation}
\mathcal{L} = \sum_{(i,j)} w_{ij} \cdot (\hat{s}_{ij} - s_{ij})^2, \quad \text{where} \quad w_{ij} = \exp(2 \cdot s_{ij})    
\end{equation}
This weighting scheme acts as a powerful form of hard-positive mining. It strengthens gradients for semantically close pairs, encouraging the model to learn subtle distinctions rather than just separate dissimilar images. This is complemented by a dataloader strategy that explicitly oversamples hard-positive pairs (details in Appendix \ref{appendix-dataloader}).

\textbf{Inference Pipeline. } During inference, for a query image/graph pair ($I_q$, $G_q$), we first pass its objects and triplets through the trained IPM to acquire $\hat{s}(u)$. These scores are used to generate the pruned scene graph $G_q^p$. $G_q^p$ and the global visual features of $I_q$, are then fed into the trained dual-stream encoder to produce the final multimodal embedding $E_q^M$. Retrieval is performed by ranking the inner product similarity $\hat{s}_{qk} = \langle E_q^M, E_k^M \rangle$ against a pre-computed candidate gallery. Because the pipeline maps each input to a single expressive multimodal embedding, candidate vectors can be pre-indexed into standard vector search engines (e.g., FAISS, HNSW) to achieve low-latency retrieval at scale.
\section{Experiments}
\label{sec:experiments}

\begin{table*}[t]
\centering
\caption{Comparison across methods on SG retrieval benchmarks. Left: PSG, Right: Automatically generated data with  PSGTR and BLIP on Flickr dataset. Best per column in \textbf{bold}.}
\label{tab:combined_datasets}
\resizebox{\textwidth}{!}{
\begin{tabular}{l|lccccccc|ccccccc}
\toprule
& \multirow{2}{*}[-1.0\normalbaselineskip]{\textbf{Method}} 
& \multicolumn{7}{c|}{\textbf{PSG Dataset}}
& \multicolumn{7}{c}{\textbf{Flickr Dataset}} \\
\cmidrule(lr){3-9} \cmidrule(lr){10-16}
& & \multicolumn{3}{c}{\textbf{NDCG}} & \multicolumn{3}{c}{\textbf{MAP}} & \textbf{MRR}
& \multicolumn{3}{c}{\textbf{NDCG}} & \multicolumn{3}{c}{\textbf{MAP}} & \textbf{MRR} \\
\cmidrule(lr){3-5} \cmidrule(lr){6-8} \cmidrule(lr){9-9} 
\cmidrule(lr){10-12} \cmidrule(lr){13-15} \cmidrule(lr){16-16}
& & @1 & @3 & @5 & @1 & @3 & @5 & -
& @1 & @3 & @5 & @1 & @3 & @5 & - \\
\midrule

\multirow{2}{*}{\rotatebox[origin=c]{90}{\textbf{Vision}}}
& DINOv3         & 78.27 & 78.35 & 78.56 & 73.81 & 80.41 & 78.98 & 83.00
& 50.84 & 49.56 & 49.22 & 55.10 & 64.32 & 63.19 & 68.33 \\
\addlinespace[2pt]
& MambaVision    & 75.32 & 75.77 & 76.18 & 68.88 & 76.28 & 75.46 & 79.14
& 50.32 & 50.80 & 50.78 & 56.20 & 64.20 & 60.24 & 68.76 \\
\midrule

\multirow{2}{*}{\rotatebox[origin=c]{90}{\textbf{VL}}} 
& BLIP-2      & 80.66 & 80.86 & 80.90 & 77.47 & 83.78 & 82.25 & 85.81
& 52.34 & 52.17 & 51.87 & 60.23 & \textbf{68.92} & 67.57 & 72.57 \\
& CLIP        & 81.28 & 81.49 & 81.59 & 78.96 & 85.12 & 83.64 & 86.90
& 52.47 & 52.59 & 52.40 & 60.59 & 68.60 & 67.79 & 72.59 \\
\midrule

\multirow{4}{*}{\rotatebox[origin=c]{90}{\textbf{Graph}}}
& Hi-SIGIR     & 81.24 & 81.68 & 81.92 & 62.40 & 52.81 & 47.68 & 74.44
& 56.49 & 57.36 & 59.27 & 50.36 & 52.59 & 47.52 & 56.15 \\
& IRSGS        & 71.39 & 72.81 & 73.28 & 60.60 & 70.08 & 69.60 & 73.11
& 30.35 & 31.43 & 31.73 & 28.75 & 37.42 & 38.67 & 43.77\\
& SCENIR       & 51.41 & 48.20 & 46.88 & 32.01 & 21.33 & 16.90 & 44.01
& 28.41 & 28.84 & 29.14 & 25.52 & 18.55 & 15.28 & 40.42 \\
& \textbf{PRISm} & \textbf{88.54} & \textbf{89.24} & \textbf{89.70} & \textbf{91.62} & \textbf{94.25} & \textbf{93.25} & \textbf{95.09}
& \textbf{60.01} & \textbf{62.84} & \textbf{64.08} & \textbf{61.25} & 67.25 & \textbf{70.15} & \textbf{74.35} \\

\bottomrule
\end{tabular}}
\end{table*}

\textbf{Datasets and Metrics.} 
We evaluate PRISm on two datasets. The main benchmark is the Panoptic Scene Graph (PSG) dataset \cite{yang2022panoptic}, which has become the standard benchmark for SG-based retrieval, 
\begin{wraptable}{r}{0.51\columnwidth}
\vspace{-5pt}
\centering
\caption{SG statistics: PSG vs.\ Flickr with generated SGs.
Edges are directed; density and degree are undirected.}
\label{tab:psg_flickr_stats}
\begin{tabular}{@{}lrr@{}}
\toprule
 & \textbf{PSG} & \textbf{Flickr} \\
\midrule
Graphs          & 47{,}874  & 12{,}000  \\
Nodes           & 535{,}441 & 139{,}577 \\
Edges           & 268{,}114 & 139{,}811 \\
Obj.\ classes   & 133       & 132       \\
Rel.\ types     & 56        & 48        \\
Nodes / graph   & 11.18     & 11.63     \\
Edges / graph   & 5.60      & 11.65     \\
Density         & 0.239     & 0.305     \\
Avg.\ degree    & 1.00      & 1.79      \\
\bottomrule
\end{tabular}
\vspace{-8pt}
\end{wraptable}
offering a more curated and standardized version of earlier datasets such as Visual Genome \cite{krishna2016visualgenome} and GQA \cite{hudson2019gqa}. 
To emulate a real-world web search setting, in which neither captions nor
SG annotations are available, we follow \cite{chaidos2025scenir} and
employ Flickr \cite{flickr30k}, discarding its human-written captions
and instead generating SGs and captions automatically with PSGTR
\cite{yang2022panoptic} and BLIP-Base \cite{blip}, respectively. We select
PSGTR for its competitive accuracy and fast inference as a one-stage,
end-to-end framework \cite{psgtr_comparison_1, psgtr_comparison_2,
ozsoy2025location}.
As summarized in Table~\ref{tab:psg_flickr_stats}, the resulting
machine-generated graphs are considerably denser than the PSG ground
truth, containing roughly twice as many edges per graph (11.65 vs.\ 5.60)
for a comparable number of nodes (11.63 vs.\ 11.18). This noisier setting
therefore stress-tests the Importance Prediction Module (IPM) in its
capacity to prune spurious relations from uncurated graphs.
The 47,874 PSG and 12,000 Flickr30k images are each randomly split into
80\% training, 12\% validation and 8\% test subsets.
For evaluation, each test image serves as a query, and the remaining test images are ranked by similarity to the query. 
We evaluate using standard retrieval metrics: Normalized Discounted Cumulative Gain (NDCG@k), Mean Average Precision (MAP@k), and Mean Reciprocal Rank (MRR).

\textbf{Baselines. } We compare PRISm against several state-of-the-art (SotA) multimodal architectures: three graph-based retrieval frameworks (IRSGS \cite{yoon2021imagegraph}, SCENIR \cite{chaidos2025scenir} and Hi-SIGIR \cite{wang2023hisigir}), two vision models (DINOv3\cite{Simeoni2025DINOv3} and Mamba Vision\cite{Hatamizadeh2024MambaVisionAH}), and two vision-language (VL) models (BLIP-2\cite{blip2} and CLIP\cite{radford2021clip}). 
To ensure a fair and rigorous comparison, all baselines, including SCENIR, were evaluated on our own standardized train/test splits. 
For all models, we use the final-layer representations as embeddings to compute similarity scores between query and candidate images.


\textbf{Implementation Details.} The IPM uses a 3-layer Transformer encoder (hidden size 1536, 32 heads) followed by 4 learned queries, with importance threshold $b$=0.4. The retrieval GNN has 3 layers (Edge-Aware layer first, residual connections on layers 2 and 3), and the global visual embedding is normalized to $\alpha$=0.7. The retrieval model is trained for 60 epochs with Adam, a learning rate of $10^{-4}$ held for 20 epochs and then decayed by $\gamma$=0.9 per epoch, batch size 32, and dropout $p$=0.1 (Appendix~\ref{appendix-details}).

\subsection{Quantitative Results}

Quantitative results on PSG in Table \ref{tab:combined_datasets} show that VL models perform markedly better than pure vision-only baselines, underscoring the necessity of multimodal reasoning for semantic-based retrieval. Moreover, PRISm significantly surpasses prior GNN-based models, specifically IRSGS and SCENIR, which rely purely on the graph modality. PRISm achieves consistently superior performance across all retrieval metrics. It substantially outperforms CLIP, the second strongest baseline, corresponding to an absolute gain in NDCG@1 of +$7.3\%$, MAP@1 +$12.7\%$, and MRR +$8.2\%$. The largest relative improvement of +$16\%$ on MAP@1 highlights PRISm’s strong ability to retrieve the most relevant item at the very top of the ranking. 

For the \textit{in-the-wild} setting on Flickr (Table~\ref{tab:combined_datasets}), where both SGs and captions are machine-generated, PRISm remains best on all top-1 metrics and MRR, improving NDCG@1, MAP@1 and MRR over the strongest baseline per metric by 3.5, 0.7 and 1.8 points, respectively, while BLIP-2 and CLIP are only slightly ahead on MAP@3. Scores drop for all methods relative to PSG. Because PSGTR is trained end-to-end on MSCOCO, its bounding-box and scene graph errors on Flickr are coupled; yet, as Table~\ref{tab:psg_flickr_stats} shows, the generated graphs carry about twice as many edges as ground-truth data and are out-of-distribution, pointing to noisy topology as the main driver of the drop. 
PRISm’s lead at the top of the ranking under this noise indicates robustness to imperfect SGs,
a prerequisite for deployment on unannotated, real-world Web imagery.

\begin{figure}[t]
    \centering
    \includegraphics[width=\linewidth]{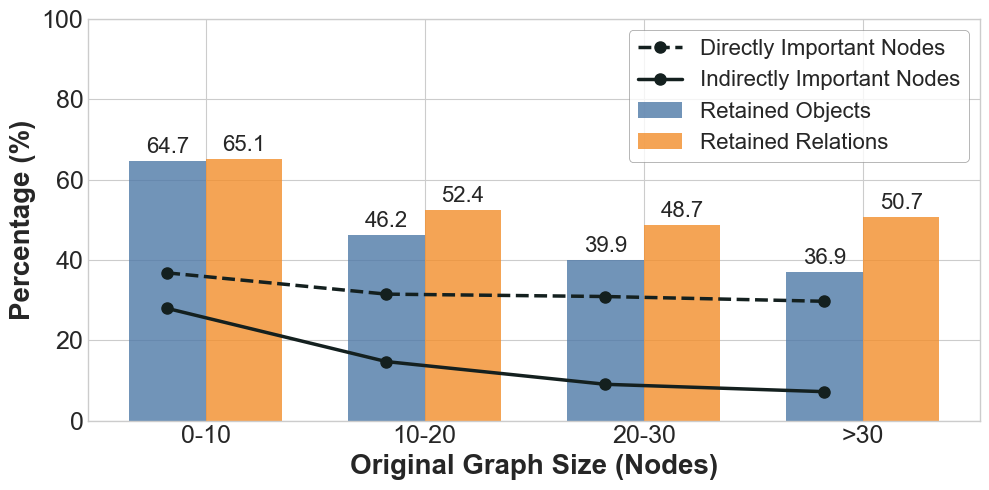}
    \caption{Object and relation retention rates by graph size.}
    \label{fig:importance_module_metrics}
    \vspace{-10pt}
\end{figure}

We further evaluate the \textbf{Importance Prediction} module as a standalone binary classifier. On the test set (without access to ground-truth captions) the module achieves a Recall of $92.54\%$ and an F1-score of $92.98\%$, showing high accuracy in distinguishing salient objects and relations. Figure \ref{fig:importance_module_metrics} illustrates the pruning behavior relative to SG complexity: as the original graph size increases, the percentage of retained objects drops significantly, from $64.7\%$ in small SGs to $36.9\%$ in large SGs. Notably, the proportion of \textit{'Directly Important'} nodes remains stable across sizes, whereas \textit{'Indirectly Important'} nodes (retained due to an important triplet) diminish in larger graphs. This confirms the IPM filters out increasing noise inherent in larger SGs while consistently preserving core semantics.

\begin{figure*}[ht]
    \centering

    \begin{subfigure}{\linewidth}
        \centering
        \includegraphics[width=\linewidth]{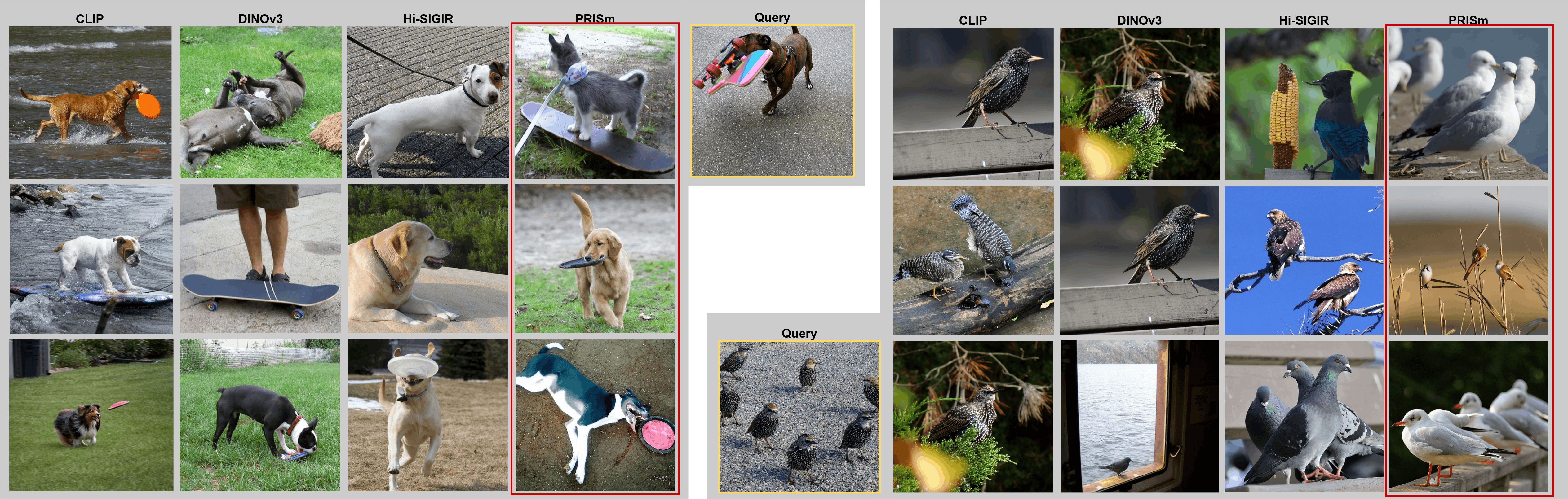}
        \caption{}
        \label{fig:main-qual}
    \end{subfigure}

    \begin{subfigure}{\linewidth}
        \centering
        \includegraphics[width=0.98\linewidth]{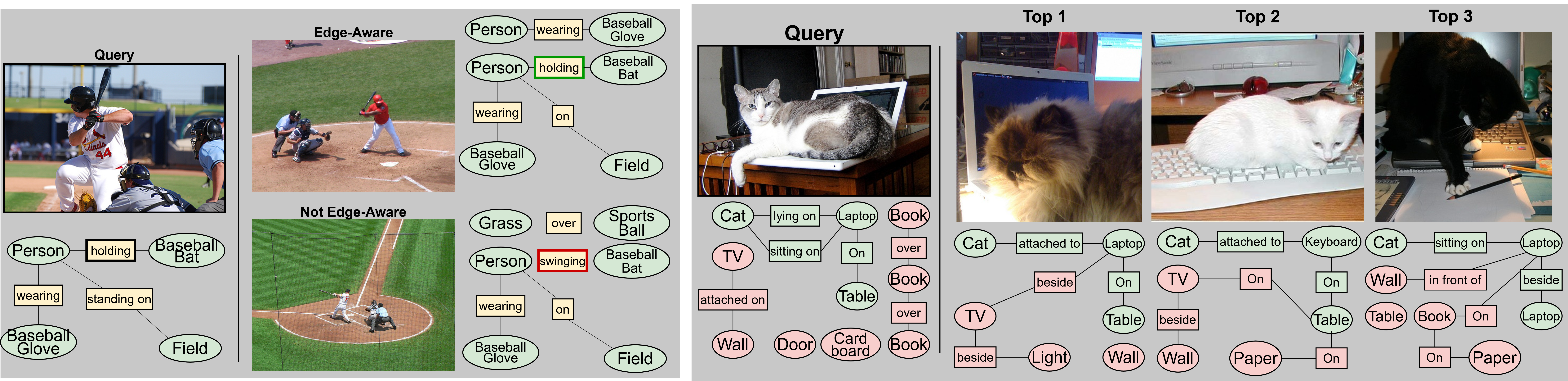}
        \caption{}
        \label{fig:qual-with-graphs}
    \end{subfigure}

    \caption{
        Retrieval examples from PRISm.  
        (a) Top-3 retrieval comparisons against best-performing SotA methods (CLIP (VL), DINOv3 (Vision), and Hi-SIGIR (Scene-Graph-based)).  
        (b) Additional PRISm retrievals illustrating detail-focused matching via the Edge-Aware GNN, as well as the effects of importance pruning and the synergy between graph semantics and visual cues.
    }
    \label{fig:merged-qual}
    \vspace{-5pt}
\end{figure*}



\textbf{Ablation Studies.} Table~\ref{tab:one_table_ndcg_map_mrr_slim}
details the ablation analysis on PSG, where human-annotated SGs and
captions provide clean supervision for each component. On Flickr, both
importance and similarity are derived from generated captions, so
differences between variants would additionally reflect errors of
external generators.
The ablation highlights a clear division of labor. The Global Visual
Stream acts as a coarse retriever, and its removal lowers NDCG@1 by 5.33
points. Notably, however, the resulting variant, which still uses
object-level visual features within the graph, outperforms every
baseline in Table~\ref{tab:combined_datasets} on all three metrics (e.g., 82.09 vs.\
78.96 MAP@1 for CLIP, the strongest baseline). The graph stream is thus
competitive on its own and governs precision ranking on hard,
relation-sensitive queries, with the global stream complementing rather
than carrying it.
When isolating the graph pipeline from visual input
($\text{PRISm}^*$, bottom tier), the Importance Prediction Module
contributes $+3.00$ points in MAP@1 and $+1.64$ in NDCG@1, confirming its
ability to suppress uninformative graph elements.
In the joint multimodal setting, where performance on curated PSG graphs
is already high (88.54 NDCG@1), the IPM and the Edge-Aware layer yield
smaller but consistent gains across all three metrics (up to $+1.05$
MAP@1 for the IPM), which are most visible on fine-grained object
interactions (Figure~\ref{fig:qual-with-graphs}).

\begin{figure}[b]
    \vspace{-5pt}
    \centering
    \includegraphics[width=0.85\linewidth]{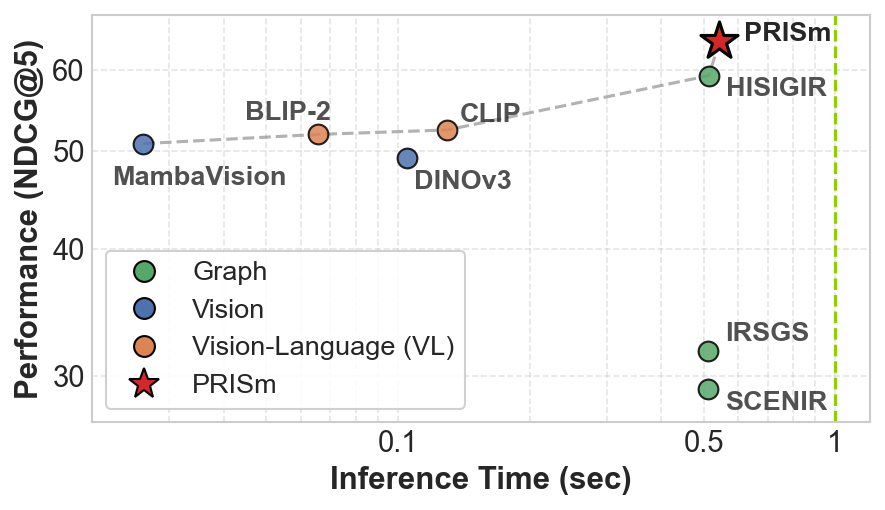}
    \caption{Inference time versus Performance of  retrieval frameworks on Flickr.}
    \label{fig:inference_times}
    \vspace{-10pt}
\end{figure}

\textbf{Efficiency and Deployability.} In Web search, gallery embeddings are computed offline and served from an approximate nearest-neighbor index, so the relevant cost is the online path of a single query. For PRISm this path consists of scene graph generation (SGG), IPM scoring, dual-stream encoding, and one nearest-neighbor lookup. The PRISm-specific components are lightweight: the IPM scores each object and triplet as an independent five-token sequence, so its cost grows linearly with graph size and all items are scored in one batched pass, while the Edge-Aware layer adds a single MLP evaluation per edge, in the first GNN layer only, leaving the linear complexity of GATv2 unchanged. On a T4 GPU, PRISm processes a query in 33\,ms, plus 510\,ms for PSGTR; SGG therefore accounts for about 94\% of query latency, and the gap to single-encoder baselines (Figure \ref{fig:inference_times}) shrinks directly with faster scene graph generators. Because each image is mapped to a single vector, candidates are pre-indexed once, and each query costs one encoding plus a sublinear lookup in a standard index (e.g., FAISS, HNSW). Matching-based methods such as Hi-SIGIR, in contrast, score every query-candidate pair with a forward pass, so their per-query cost grows linearly with gallery size and cannot be amortized by indexing. In exchange for its SGG overhead, PRISm improves NDCG@5 by 22.3\% (relative) over the strongest vision/VL model and by 8.1\% over the strongest graph-based method.





\begin{table}[t]
\centering
\vspace{-5pt}
\caption{PRISm ablations on PSG. PRISm* is  PRISm without any visual information. Best per column and among variants in \textbf{bold}.}
\label{tab:one_table_ndcg_map_mrr_slim}
\small 
\setlength{\tabcolsep}{3.5pt} 
\begin{tabular}{@{}l|lccc@{}} 
\toprule
& \textbf{Setting} & \textbf{NDCG@1} & \textbf{MAP@1} & \textbf{MRR} \\
\midrule

\multirow{4}{*}{\rotatebox[origin=c]{90}{\makecell{\textbf{With}\\\textbf{Vision}}}}
& PRISm                   & \textbf{88.54} & \textbf{91.62} & \textbf{95.09} \\
& - Importance Pruning    & 88.05          & 90.57          & 94.40 \\
& - Edge-Aware Layer      & 88.39          & 91.46          & 95.04 \\
& - Global Visual Stream  & 83.21          & 82.09          & 88.70 \\
\midrule

\multirow{3}{*}{\rotatebox[origin=c]{90}{\makecell{\textbf{Without}\\\textbf{Vision}}}}
& PRISm*                  & \textbf{74.09} & \textbf{64.93} & \textbf{76.84} \\
& - Edge-Aware Layer      & 73.70          & 64.15          & 76.01 \\
& - Importance Pruning    & 72.45          & 61.93          & 74.24 \\
\bottomrule
\end{tabular}
\vspace{-5pt}
\end{table}

\subsection{Qualitative Results}
\label{sec:qual}

Firstly, in Figure~\ref{fig:main-qual}, we present the top-3 retrievals from the best-performing models of each category: CLIP (VL), DINOv3 (visual), Hi-SIGIR (SG-based), and our model, PRISm. The left example corresponds to a query of a dog biting a skateboard. While most models correctly identify that similar images should include dogs interacting with objects, they struggle with the specific and atypical relation: a dog interacting with a skateboard, an object more commonly associated with humans and the relation \textit{"riding"}. PRISm is the only model that retrieves an image containing both key objects, the dog and the skateboard. The second and third retrievals from PRISm, although missing the skateboard, accurately preserve the \textit{"dog-biting"} interaction. Notably, PRISm’s top-2 result also shows the dog in a pose highly consistent with the query, outperforming visually similar \textit{“dog-biting-frisbee”} results retrieved by CLIP (top-1), DINOv3 (top-3), and Hi-SIGIR (top-3). Other models often miss meaningful interactions and sometimes even miss the dog itself (e.g., DINOv3 top-2).  Overall, the closest competitor visually is the VL model, followed by the SG-based one, supporting our findings in Table~\ref{tab:combined_datasets} (left) and underscoring the importance of integrating both visual and semantic reasoning for accurate image retrieval.

Regarding Fig.~\ref{fig:main-qual} (right), PRISm shows strong ability in retrieving images with relevant objects and \textit{also} preserving their multiplicity. The query image depicts seven birds, and PRISm successfully retrieves scenes with several birds, capturing both object presence and relational context. In contrast, models relying more heavily on cues such as color (CLIP, DINO) tend to retrieve single-bird images, likely due to superficial visual similarity. The graph-based Hi-SIGIR performs better in maintaining object multiplicity but still fails to capture it in the most important top result.


Additionally, Fig.~\ref{fig:qual-with-graphs} (left) demonstrates the critical role of our Edge-Aware GNN in distinguishing fine-grained interactions. For a query of a baseball player in a batting stance, a standard GNN 
focuses on object co-occurrence, incorrectly retrieving a scene characterized by a \textit{"swinging"} relation. In contrast, PRISm’s Edge-Aware GNN explicitly encodes the structural context, correctly identifying the \textit{"holding"} interaction and retrieving a visually and semantically congruent image. Furthermore, Fig.~\ref{fig:qual-with-graphs} (right) highlights the model’s ability to perform semantic retrieval through importance pruning. For a cluttered scene of a cat on a laptop, PRISm retains the key \textit{"Cat-sitting on-Laptop"} triplet while discarding less relevant nodes (e.g., \textit{"Book"}, \textit{"TV"}). This focus enables the retrieval of top-ranked images that strictly adhere to this core semantic concept (regardless of variations in the background) demonstrating a retrieval process that prioritizes semantic relevance over superficial visual matching.

\section{Conclusion}
\label{sec:conclusion}

We presented PRISm, an image-to-image retrieval framework that prunes
scene graphs to their semantically salient elements and fuses them,
through relation-aware message passing, with global visual context into a
single index-ready embedding. Beyond its gains over vision,
vision-language and SG-based baselines on both curated and
machine-generated graphs, our analysis yields three broader observations.
First, explicit relational structure is competitive on its own: even
without the global visual stream, PRISm outperforms all baselines on PSG.
Second, the IPM prunes more aggressively as graphs grow, suggesting that
learned importance matters most where generated graphs are densest.
Third, scene graph generation, rather than the retrieval model, dominates
query latency, so faster generators translate directly into cheaper
structure-aware search. Future work includes end-to-end adaptation of the
frozen encoders and evaluation on larger, Web-scale galleries.


\bibliographystyle{ACM-Reference-Format}
\bibliography{references}

\appendix

\section{Method Details}

\subsection{Unsupervised Graph Representation with InfoGraph}
\label{appendix-infograph}
The IPM receives an unsupervised graph embedding $\mathbf{z}_G$ from InfoGraph~\cite{sun2020infograph}, so that importance can depend on the structural context of each element and not only on its label: a horse, for instance, matters more when a person is riding it than when it appears alone. We convert each scene graph into a bipartite graph in which every relation becomes its own node, connected to the two objects it links, and keep only textual node features; visual information is not used at this stage. InfoGraph is fully unsupervised, scales to large graph collections, and captures global structure by maximizing mutual information between local and global representations. We train it for 150 epochs on the bipartite training graphs, then apply the trained encoder in a single forward pass to both training and test graphs to obtain the fixed-dimensional embeddings fed to the IPM.

\subsection{Binary Clustering with Jenks}
\label{appendix-jenks}
For each image, the IPM produces two score lists, one for objects and one for triplets, which are binarized separately since triplets carry more information than single objects. A single global threshold proved unsuitable because score distributions vary widely across images: a fixed cutoff can eliminate every element in one image while a low value (e.g., 0.4) suffices in another, and many images mix very high and very low scores. We therefore complement the absolute threshold $b$ with the Jenks natural breaks algorithm~\cite{jenks1967data}, which partitions each list into two classes by minimizing within-class and maximizing between-class variance, separating ``high'' from ``low'' importance adaptively for each image.

\subsection{Hard-Positive Sampling and Loss Weighting}
\label{appendix-dataloader}

\begin{figure}
    \centering
    \includegraphics[width=0.85\linewidth]{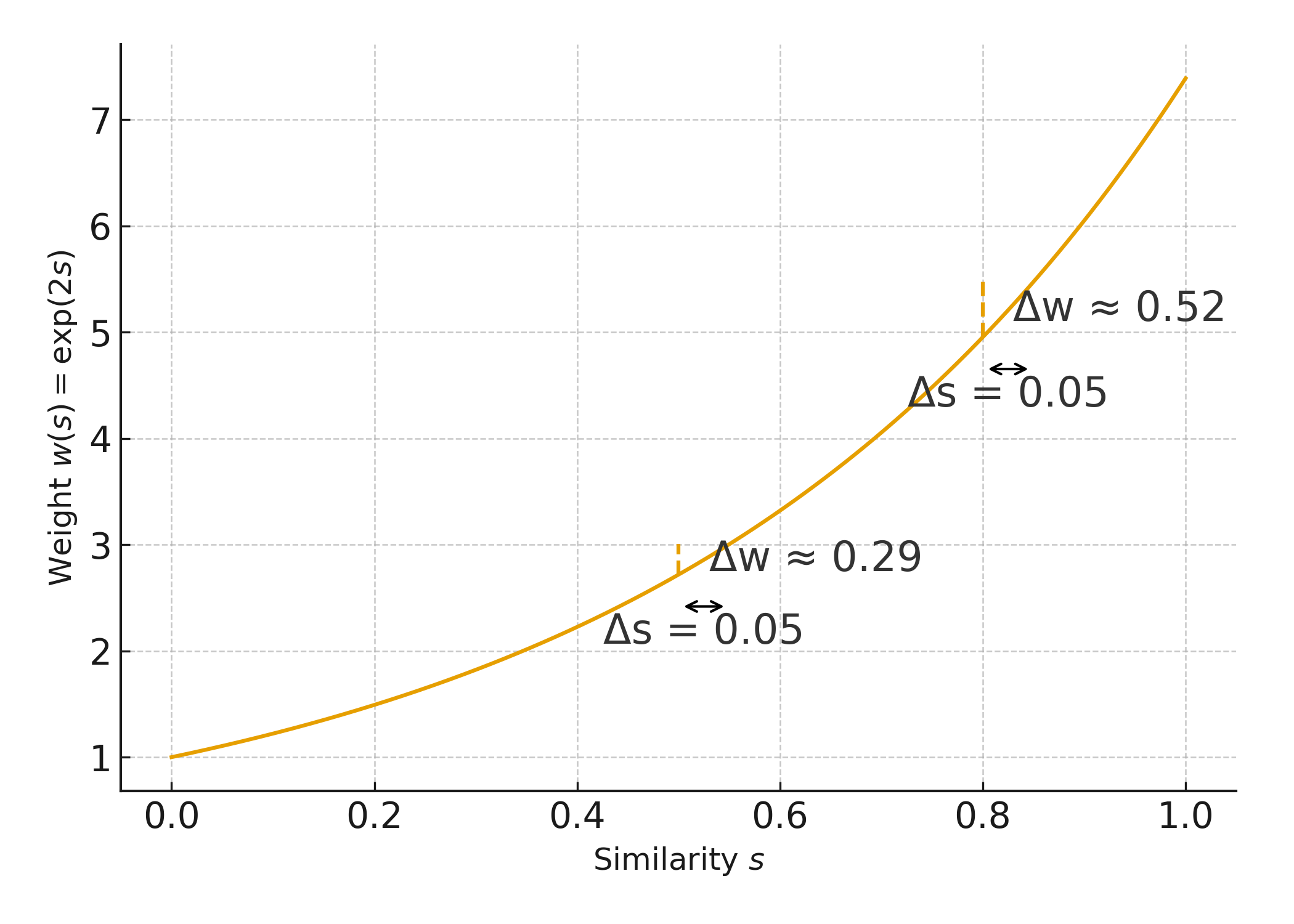}

    \caption{
    Effect of the weighted loss function $w(s) = \exp(2s)$ used when training the retrieval model to favor highly similar pairs.
        }
    \label{fig:exp_weight_function}
    \vspace{-10pt}
\end{figure}

Retrieval requires distinguishing highly similar images, which often share overlapping object sets and differ only slightly in ground-truth similarity. We emphasize such pairs with two complementary techniques. First, the dataloader oversamples hard positives: with probability $0.05$, the partner of a training query image is drawn from its top-5 most similar images (by ground-truth similarity) instead of uniformly. Second, the loss of each pair is weighted by $w(s) = \exp(2s)$ (Figure~\ref{fig:exp_weight_function}). Because the exponential grows faster at high similarity, the same difference $\Delta s = 0.05$ changes the weight noticeably more near $s = 0.80$ than near $s = 0.50$; gradients therefore concentrate on fine distinctions among the most similar pairs, while low- and medium-similarity pairs, whose exact scores matter little for retrieval, contribute relatively less.

\begin{table}[b]
\centering
\vspace{-5pt}
\caption{Effect of oversampling and weighted loss on top-1 retrieval performance. The best result per column is denoted in \textbf{bold}.}
\begin{tabular}{lccc}
\hline
Setting              & NDCG@1 & MAP@1 & MRR   \\
\hline
PRISm                & \textbf{88.54}  & \textbf{91.62} & \textbf{95.09} \\
No oversampling      & 87.79  & 90.81 & 94.45 \\
No weighted loss     & 88.02  & 91.25 & 94.74 \\
\hline
\end{tabular}
\label{tab:ndcg1_map1_mrr}
\end{table}

Table~\ref{tab:ndcg1_map1_mrr} ablates both components: removing either one lowers all top-1 metrics, with oversampling having the larger effect.

\section{Experiments}

\subsection{Further Implementation Details}
\label{appendix-details}

To instantiate the modality encoders $e(\cdot)$ and $m(\cdot)$ used strictly during the training and inference phases of PRISm, we employ the OpenCLIP framework \citep{ilharco_gabriel_2021_5143773}. Specifically, we extract visual and textual representations using the OpenCLIP Vision Transformer and transformer-based text encoder, respectively. Conversely, to compute the ground-truth similarity scores, we independently utilize the \texttt{all-mpnet-base-v2} Sentence-BERT (SBERT) model \citep{sbert}. Decoupling the ground-truth generation from the model's feature extractors eliminates representation conflict and prevents information leakage between the target supervision signal and the learned embeddings.
As for the data partitions, since papers like SCENIR do not offer publicly available splits, we fully reproduce their framework on our own standardized splits, naturally accounting for any metric variation compared to the original paper.
Our GNN consists of three layers with hidden
dimensions [6144, 2048, 2048]. The global visual embeddings are processed by a
two-layer MLP with dimensions [1024, 1024], using ReLU activation between the
layers and including bias terms throughout.

For the Importance Prediction module, after the multi-head attention operation
with learned queries, we applied mean pooling over the resulting token
representations and passed the pooled embedding through a 3-layer MLP with
dimensions $[1536,\, 768,\, 1]$ and GELU activations to produce the final scalar
importance score. Each Transformer encoder layer employed a dropout rate of
$p=0.1$ to mitigate overfitting. The Importance Prediction module was trained for 10 epochs with an initial learning rate of
$10^{-5}$, using a learning-rate scheduler consistent with that of the retrieval
GNN: the learning rate was reduced by 10\% after every epoch, starting at epoch~1.
Training was performed with a batch size of 32, and early stopping with a patience
of 3 epochs was applied based on validation-loss improvements to prevent
overfitting. Notably, our entire pipeline was trained on a single L40s GPU (48GB VRAM) in $\sim$2.5 hours.

For all remaining baselines, we leverage their official implementations (with the exception of Hi-SIGIR, which we re-implement following the original paper due to the absence of publicly available code) and tune their architectural hyperparameters separately on the PSG and Flickr datasets.

To evaluate inference latency on the Flickr test split (Figure~\ref{fig:inference_times}), we use a batch size of 1 (matching a single query against the 959 remaining candidate images) and average the runtime across all 960 test queries. For each query, the timed region encompasses scene graph generation (when required by the method), the model's core submodules (IPM scoring and dual-stream encoding for PRISm), and the final retrieval lookup. Since Hi-SIGIR is a matching-based retrieval model (i.e. query-candidate pair input), the latency was measured across all possible pairs of the same 960-image gallery.


\subsection{Retrieval Implementation for Vision and Vision-Language Models}
For all vision and vision-language (VL) baselines, we use only the visual encoder to extract a global image descriptor. Images are loaded in RGB and preprocessed with each model's official evaluation transforms (resize $\rightarrow$ crop $\rightarrow$ normalize); embeddings are computed in batches with gradients disabled, $\ell_2$-normalized so that cosine similarity reduces to an inner product, and stacked into an $N \times D$ matrix for retrieval. The descriptor is the output of the visual transformer for OpenCLIP~\cite{ilharco_gabriel_2021_5143773} (\textit{ViT-H/14}, \textit{laion2b\_s32b\_b79k}); the pooled output of the vision model, with pixel tensors built by the \textit{Blip2Processor}, for BLIP-2~\cite{blip2} (\textit{Salesforce/blip2-opt-2.7b}); the first [CLS] token of the final layer for DINOv3~\cite{Simeoni2025DINOv3} (\textit{dinov3-vitl16-pretrain-lvd1689m}); and the average-pooled output for MambaVision~\cite{Hatamizadeh2024MambaVisionAH} (\textit{MambaVision-L-1K}), following each model's suggested usage.


\section{Further Qualitative Results}

\begin{figure}
    \centering
    \includegraphics[width=0.95\linewidth]{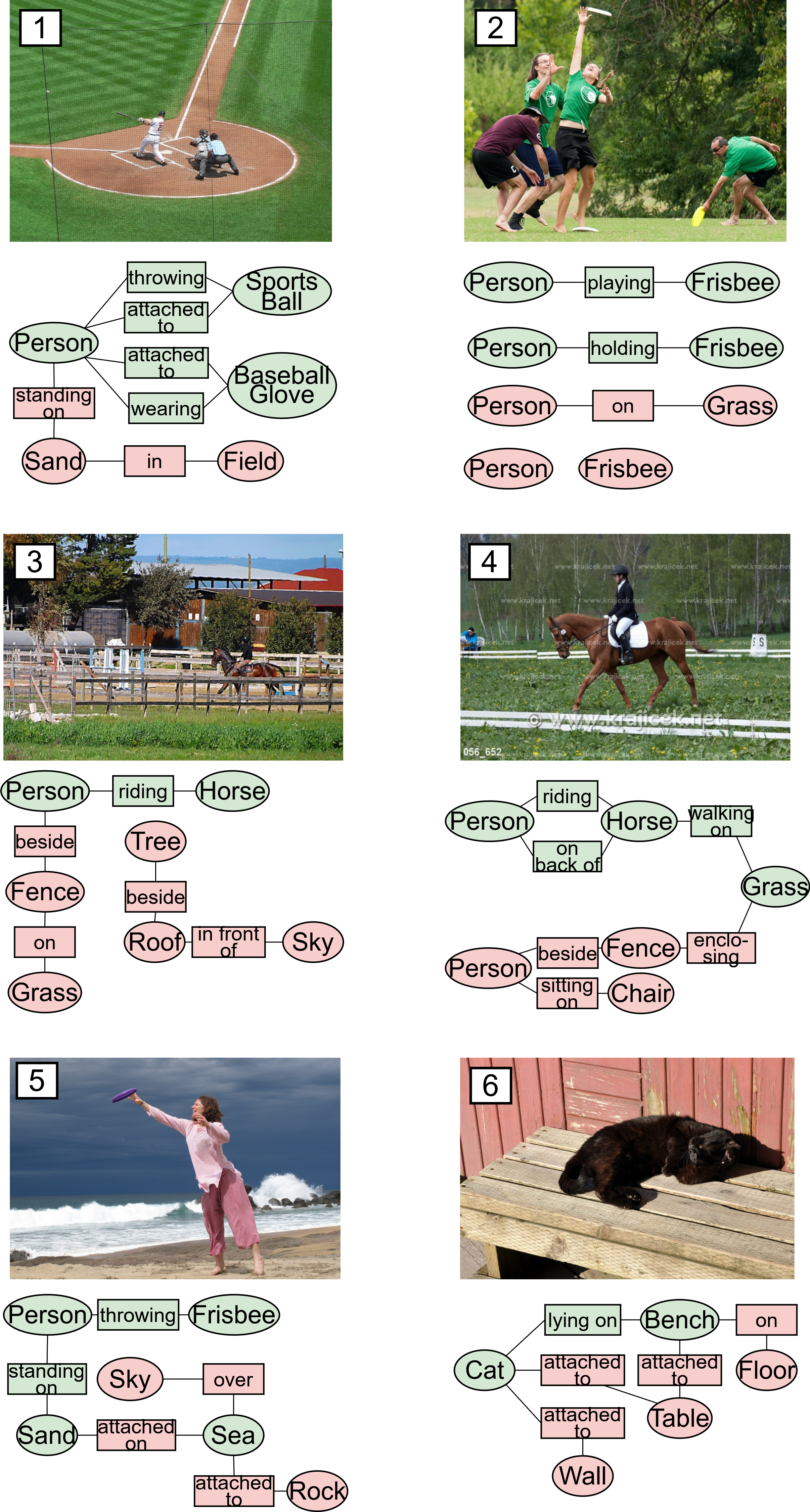}
    \caption{Qualitative outputs of the Importance Prediction Module, illustrating the selective retention of semantically critical graph elements (\textcolor{ForestGreen}{green}) and the pruning of less relevant background information (\textcolor{WildStrawberry}{red}).}
    \label{fig:importance_qual}
    \vspace{-10pt}
\end{figure}

\begin{figure*}[p]
    \centering
    \begin{subfigure}{0.495\linewidth}
        \includegraphics[width=\linewidth]{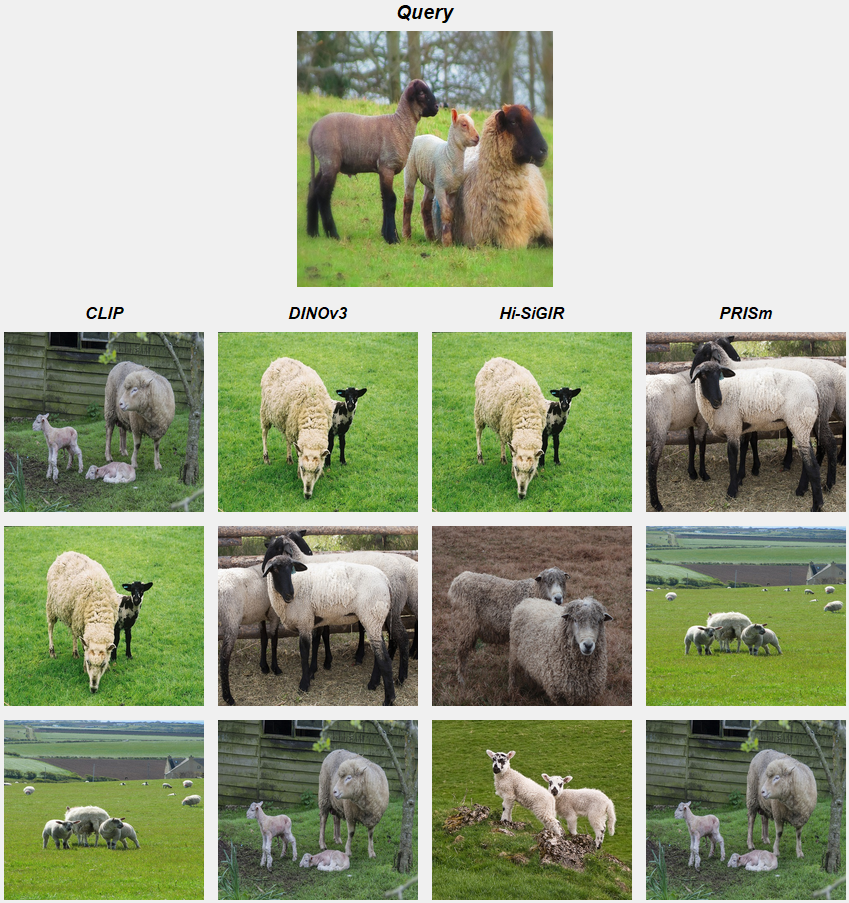}
    \end{subfigure}
    \hfill
    \begin{subfigure}{0.495\linewidth}
        \includegraphics[width=\linewidth]{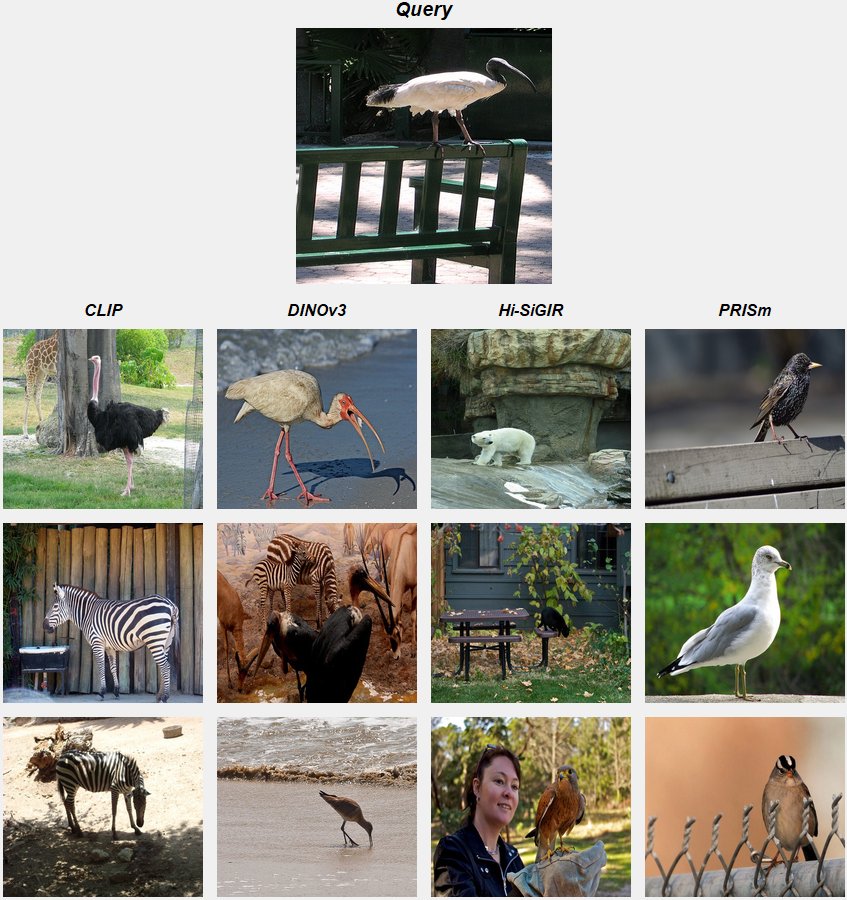}
    \end{subfigure}

    \vspace{0.2cm}

    \begin{subfigure}{0.495\linewidth}
        \includegraphics[width=\linewidth]{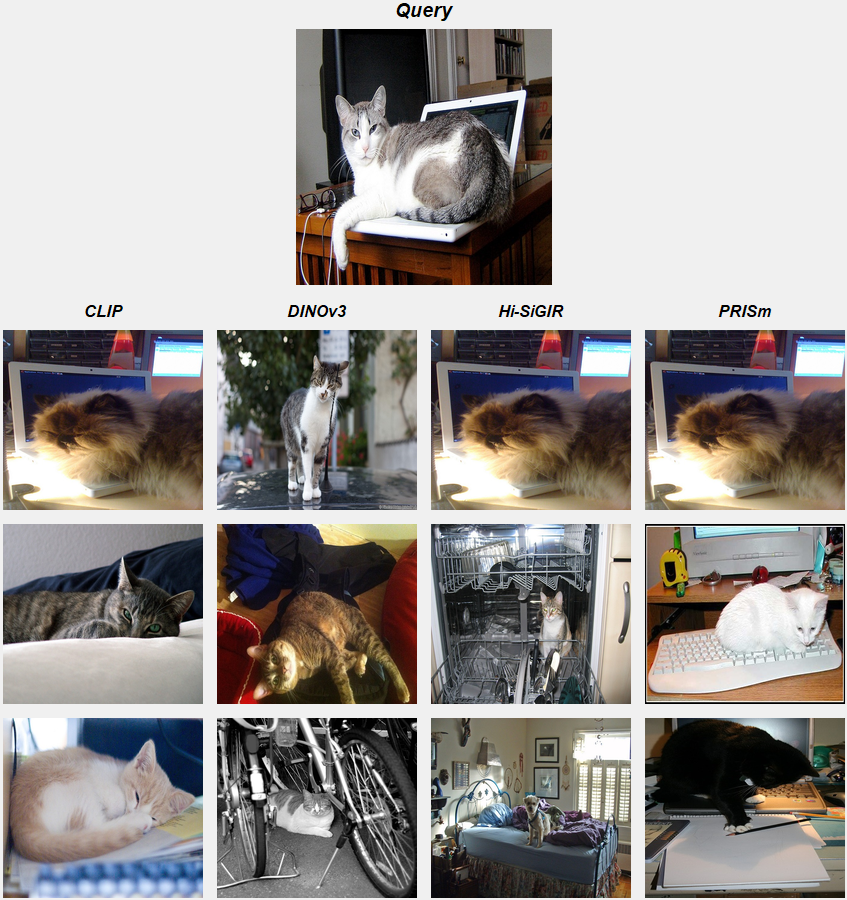}
    \end{subfigure}
    \hfill
    \begin{subfigure}{0.495\linewidth}
        \includegraphics[width=\linewidth]{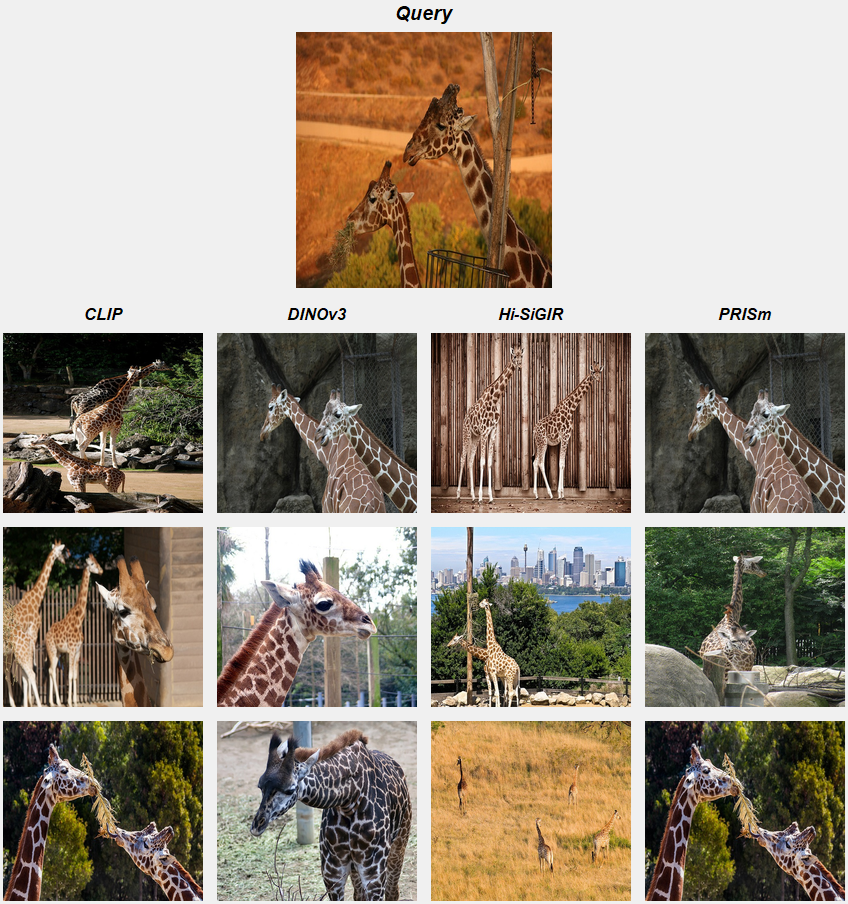}
    \end{subfigure}

    \caption{Top-3 retrieval comparisons of our proposed framework PRISm against best-performing SotA methods (CLIP (VL), DINOv3(Vision), and Hi-SIGIR (Scene-Graph-based))}
    \label{fig:qual_1}
\end{figure*}

Figure~\ref{fig:importance_qual} shows IPM outputs, with retained objects and relations in green and pruned background elements in red.
Figure~\ref{fig:qual_1} gives additional top-3 retrievals from the best model of each category: CLIP (VL), DINOv3 (vision), Hi-SIGIR (graph-based), and PRISm.

\section{Limitations}
\label{appendix:limitations}

A limitation of the PRISm framework lies in its architectural reliance on pre-trained, frozen external models. Our pipeline depends on a fixed vision encoder for extracting global and object-level visual features, a frozen sentence encoder for processing textual labels, and (in the case of unannotated "in-the-wild" data) third-party scene graph generators and captioning models. As we briefly discuss in the main text regarding the model's sensitivity to scene graph quality, the upper bound of our retrieval performance is inherently linked to the fidelity of these additional components. Consequently, noise or omissions in these fixed representations can propagate to the Importance Prediction and Edge-Aware GNN modules. However, while this modular design is essential for maintaining the high computational efficiency and low training costs highlighted in our results, it suggests that future iterations could potentially yield further improvements by exploring end-to-end fine-tuning to tighter align the backbone encoders with the specific demands of structure-aware retrieval.
Furthermore, our evaluation relies on a caption-derived surrogate of
semantic similarity rather than direct human relevance judgments. Although
this protocol is standard in SG-based retrieval \cite{yoon2021imagegraph,
wang2023hisigir}, and averaging over multiple captions mitigates the bias
of any single description, it inherits the coverage and phrasing biases
of the captions themselves and, on Flickr, of the BLIP captioner. A user
study of perceived relevance would provide a more direct assessment.
Finally, PRISm has been evaluated on galleries of up to a few thousand
images; whether it retains its advantage at larger scales, where
near-duplicates and distractors are far more frequent, remains to be
verified.

\newpage

\clearpage


\end{document}